\documentclass[journal]{IEEEtran}
\usepackage{}
\usepackage{lineno,hyperref}
\usepackage{graphicx}
\usepackage{epstopdf}
\usepackage{mathrsfs}
\usepackage{amsfonts}
\usepackage{bm}
\usepackage{color}
\usepackage{multirow}
\usepackage{threeparttable}
\usepackage[subfigure]{tocloft}
\usepackage{subfigure}
\ifCLASSINFOpdf
\else
\fi
\begin{document}

\title{Dual-Stream Collaborative Transformer for Image Captioning}

\author{Jun~Wan, ~\IEEEmembership{Member,~IEEE,} Jun~Liu, Zhihui~lai, Jie~Zhou,  ~\IEEEmembership{Member,~IEEE,}
	
	
	\thanks{ This work is supported by the National Natural Science Foundation of China (Grant No. 62571555), the Natural Science Foundation of Hubei Province, China (Grant No. 2024AFB992).}
	\thanks{J. Wan is with the School of Information Engineering, Zhongnan University of Economics and Law, Wuhan, 430073, China and (e-mail:junwan2014@whu.edu.cn).}
	\thanks{J. Liu is with the Information Systems Technology and Design
		Pillar, Singapore University of Technology and Design, Singapore, 487372 (e-mail: jun\_liu@sutd.edu.sg).}
	\thanks{J. Zhou and Z. Lai are with the Computer Vision Institute, College of Computer Science and Software Engineering, Shenzhen University, Shenzhen 518060, China, and also with the Shenzhen Institute of Artificial Intelligence and Robotics for Society, Shenzhen, China. (e-mail: jie\_jpu@163.com, lai\_zhi\_hui@163.com).}%
}

\markboth{Journal of \LaTeX\ Class Files}%
{Shell \MakeLowercase{\textit{et al.}}: Bare Demo of IEEEtran.cls for IEEE Journals}

\maketitle
\begin{abstract}
Current region feature-based image captioning methods have progressed rapidly and achieved remarkable performance. However, they are still prone to generating irrelevant descriptions due to the lack of contextual information and the over-reliance on generated partial descriptions for predicting the remaining words. In this paper, we propose a Dual-Stream Collaborative Transformer (DSCT) to address this issue by introducing the segmentation feature. The proposed DSCT consolidates and then fuses the region and segmentation features to guide the generation of caption sentences. It contains multiple Pattern-Specific Mutual Attention Encoders (PSMAEs) and Dynamic Nomination Decoders (DNDs). The PSMAE effectively highlights and consolidates the private information of two representations by querying each other. The DND dynamically searches for the most relevant learning blocks to the input textual representations and exploits the homogeneous features between the consolidated region and segmentation features to generate more accurate and descriptive caption sentences. To the best of our knowledge, this is the first study to explore how to fuse different pattern-specific features in a dynamic way to bypass their semantic inconsistencies and spatial misalignment issues for image captioning. The experimental results from popular benchmark datasets demonstrate that our DSCT outperforms the state-of-the-art image captioning models in the literature.
\end{abstract}


\IEEEpeerreviewmaketitle

\section{Introduction}
\IEEEPARstart{I}{mage} captioning aims to automatically generate descriptive sentences for a given image. It needs to identify objects, model their spatial and semantic relationships, and verbalize them using natural language. Accurate and descriptive image captioning lays the foundation of visual intelligence \cite{Chen2017DynamicallyMM, Wan2021RobustFA, Wan2021RobustAP} in chatting robots, photo sharing \cite{Amon2020InfluencingPS, Li2019HideMePP} on social media, and aids for visually impaired people \cite{Afif2020IndoorOD, Ranasinghe2016SalientFO}.
\begin{figure}[!t]
	\begin{center}
		\includegraphics[width=0.96\linewidth]{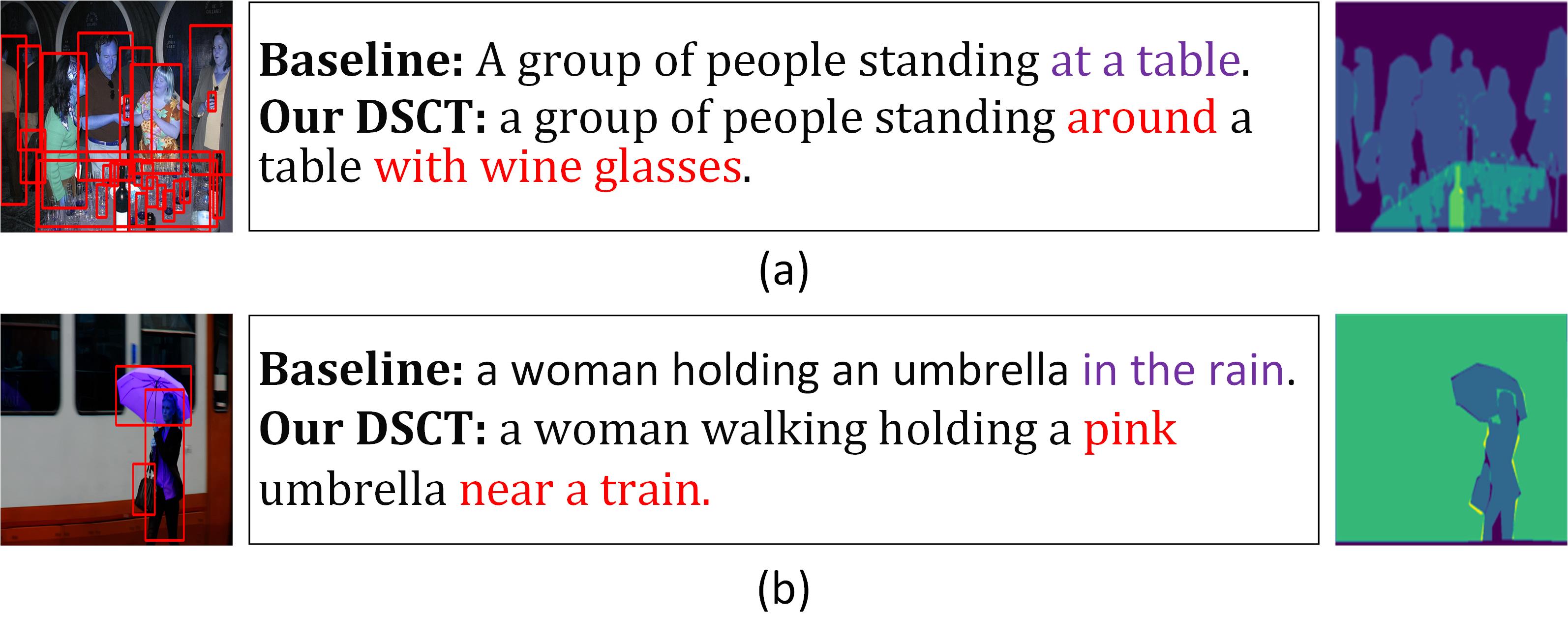}
	\end{center}
	\caption{Baseline refers to the popular transformer-based image captioning model that uses only the region feature to generate caption sentences. (a) region features lack modeling of contextual information that easily lead to inaccurate captions (e.g., \textcolor[rgb]{1,0,0}{ at a table}). (b) region feature-based captioning models highly depend on the partially generated description for predicting the remaining words, which causes irrelevant descriptions (e.g., \textcolor[rgb]{1,0,0}{in the rain}). The integration of region and segmentation features into our DSCT framework learns more effective visual and semantic representations and exploits the homogeneous features among them for more accurate and descriptive image captioning.}
	\label{problem}
\end{figure}

Most region feature-based image captioning methods \cite{Anderson2018BottomUpAT, Huang2019AttentionOA, Cornia2020MeshedMemoryTF, Pan2020XLinearAN} have achieved great success as the region feature effectively enhances the visual-semantic embeddings and can be directly used to represent and identify most of the salient objects in an image. However, the region features \cite{Ren2015FasterRT, He2016DeepRL, Krishna2016VisualGC} are extracted based on the object detection method that is originally designed to address the localization and identification of individual objects. Therefore, region features lack modeling of contextual information (e.g., the relationship between objects) that is very important for image captioning. Thus, current region feature-based captioning models fail to generate accurate image sentences (see Fig. \ref{problem} (a)). Moreover, image captioning models highly depend on partially generated descriptions, which tend to generate irrelevant caption sentences (see Fig. \ref{problem} (b)). Inspired by the fact that segmentation features \cite{WuDIFNetBV} can serve as spatial semantic guidance for inferring underlying semantics and spatial relationships, this work explores ways to enhance the visual representation and reduce the dependency on partially generated sentences by introducing segmentation features. But region and segmentation features are pattern-specific features so it is difficult to directly fuse/use them due to their semantic inconsistency and spatial misalignment. How to combine these two features to enhance visual representation and reduce the dependence on partially generated sentences for image captioning remains a challenging problem.
\begin{figure*}[!t]
	\centering
	\includegraphics[width=0.96\linewidth]{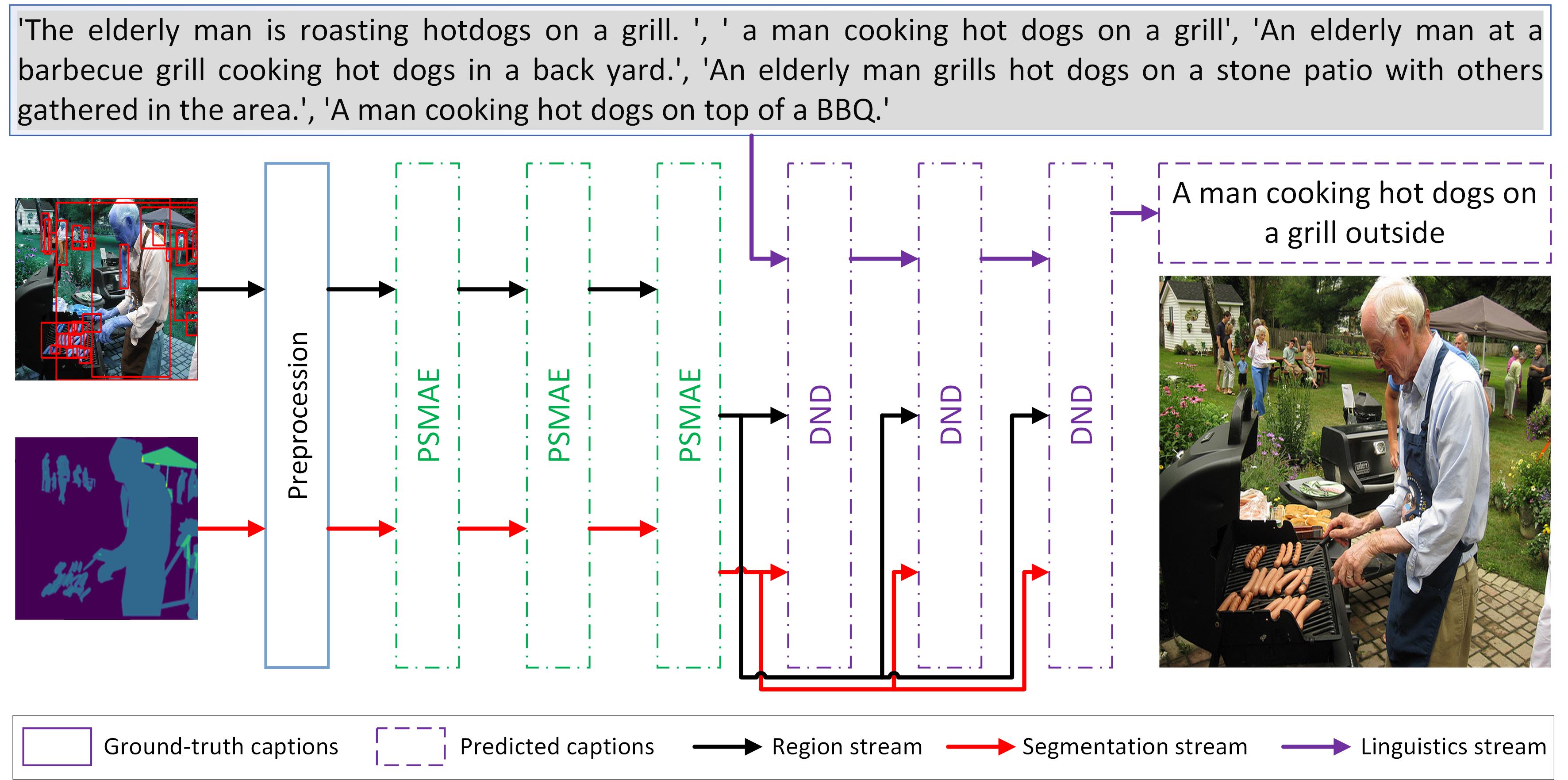}
	\centering
	\caption{The overall architecture of the proposed Dual-Stream Collaborative Transformer (DSCT). The proposed PSMAE consolidates both the region and segmentation representations by highlighting their private information and querying each other. Then, they are fused in a novel dynamic way by the proposed DND to guide the generation of caption sentences. By integrating the PSMAE and DND into a Dual-Stream Collaborative Transformer (DSCT) via a seamless formulation, the segmentation and region features are utilized more effectively for achieving more accurate and descriptive image captioning.}
	\label{dsct}
\end{figure*}

To address the above problems, in this paper, we propose a novel two-stream transformer framework, Dual-Stream Collaborative Transformer (DSCT) (as shown in Fig. \ref{dsct}) for more accurate and descriptive image captioning. It is motivated by that the function of human brain can continuously consolidate and update old memories in daily life, enabling it to search and use the most relevant memories when encountering new problems. The proposed DCST achieves this goal by combining Pattern-Specific Mutual Attention Encoder (PSMAE) and Dynamic Nomination Decoder (DND). To be specific, the proposed PSMAE is a two-stream framework that contains a region feature stream and a segmentation feature stream, which consolidates and updates both region and segmentation representations by highlighting their private information and querying each other. DND is designed to fuse consolidated region and segmentation features by dynamically searching for the most relevant learning blocks to the input textual representation and exploiting the homogeneous features between region and segmentation features to guide the generation of the next token. On the one hand, this design forces each stream to learn more private information and consolidate itself. On the other hand, the semantic inconsistency and spatial misalignment issue between the two pattern-specific features are bypassed by stacking multiple DNDs. Finally, by seamlessly integrating the proposed PSMAE and DND into a novel DSCT framework, more effective visual and semantic representations are learned to reduce the dependence on partially generated descriptions for predicting the remaining words, thereby achieving more accurate and descriptive image captioning. The main contributions of this work are summarized as follows:

1) By incorporating region feature stream and segmentation feature stream, we propose a Pattern-Specific Mutual Attention Encoder (PSMAE), in which both region and segmentation features are consolidated and updated by highlighting their private information and querying each other.

2) A Dynamic Nomination Decoder (DND) is proposed to fuse the consolidated region and segmentation features in a novel dynamic way such that their semantic inconsistency and spatial misalignment problem are bypassed and the homogeneous features between them can be effectively mined to guide the generation of caption sentences.

3) By seamlessly integrating PSMAE and DND via a two-stream framework of transformer, more effective visual and semantic representations are learned to reduce the dependence on partially generated descriptions for predicting the remaining words, and our proposed DSCT outperforms the state-of-the-art image captioning models on popular benchmark datasets COCO \cite{Lin2014MicrosoftCC}.

The rest of the paper is organized as follows. The related work and preliminaries are separately given in Section \textbf{II} and Section \textbf{III}. Section \textbf{IV} introduces the proposed DSCT including PSMAE and DND. We conduct the experiments and evaluate the proposed DSCT in Section \textbf{V}, and conclude this paper in Section \textbf{VI}.
\section{Related Work}
Many approaches have been proposed in image captioning and have achieved promising results. In general, existing captioning models can be categorized into two groups: traditional methods and deep learning methods.

\textbf{Traditional methods. } Traditional image captioning methods can be further divided into retrieval-based \cite{Farhadi2010EveryPT, Gupta2012ChoosingLO, Ordonez2011Im2TextDI} and template-based methods \cite{Kulkarni2013BabyTalkUA, Ushiku2015CommonSF}. In general, retrieval-based methods retrieve one or a set of most similar sentences from a pre-specified sentence pool. Farhadi et al. \cite{Farhadi2010EveryPT} introduce a new meaning space (denoted as a triplet $<object, action, scence>$) between the image space and the sentence space. The output image descriptions are obtained by searching sentence pools with good matching scores in this meaning space. By using information from three different sources, Gupta et al. \cite{Gupta2012ChoosingLO} propose a generic method that generates clear and semantically correct descriptions. Template-based methods first generate sentence templates with slots that are then filled with detected visual concepts, and the new description sentences can be obtained by using one or more templates. By exploiting both statistics gleaned from text data and recognition algorithms from computer vision, Kulkarni et al. \cite{Kulkarni2013BabyTalkUA} present a system that produces more relevant sentences for images. Ushiku et al. propose a novel Common Subspace for Model and Similarity (CoSMoS) \cite{Ushiku2015CommonSF}, which achieves more effective phrase learning with few training samples.

\textbf{Deep learning methods. } With the widespread of deep learning \cite{Wan2022RobustFL, Lin2022SwinBERTET, Wan2021RobustFL}, the encoder-decoder framework \cite{Xu2015ShowAA, Lu2017KnowingWT, Wan2023PreciseFL, Anderson2018BottomUpAT} has been widely exploited in captioning models. CNN and RNN are usually used as the encoder and decoder for learning visual features and generating the output descriptions, respectively. Xu et al. \cite{Xu2015ShowAA} aim to boost image captioning by integrating stochastic attention and deterministic attention, which gives more interpretability to the model generation process and learns more human-intuitive alignments. Lu et al. propose an adaptive encoder-decoder framework \cite{Lu2017KnowingWT}, which extracts meaningful information for sequential word generation by automatically deciding whether to rely on the visual information in a given image. Anderson et al. \cite{Anderson2018BottomUpAT} propose a combined bottom-up and top-down attention mechanism to improve the image captioning performance and model interpretability. Yang et al. \cite{Yang2021ExploitingCP} propose an image captioning method by exploiting the Cross-modal Prediction and Relation Consistency (CPRC). By transforming the raw image and corresponding generated sentence into the shared semantic space and then measuring the generated sentences with prediction consistency and relation consistency, CPRC can improve the captioning performance under complex scenarios. Wang et al. \cite{Wang2022GLCMGC} propose an attention-based global-local captioning model, which can model more effective correlations between words and provides more interpretability for remote sensing image captioning. Song et al. \cite{Song2022MemorialGW} address unpaired image captioning by learning the latent semantic relevance of the multimodal semantic-aware space in a novel adversarial manner. The above image captioning models fail to generate accurate and fluent caption sentences due to their inability to model the complex relationships between distant objects.

Recently, the transformer model has been introduced to address the above problem by replacing recurrence and convolutions with the attention mechanism, and has achieved remarkable performance. Herdade et al. \cite{Herdade2019ImageCT} integrate the spatial relationship between objects into the captioning model via the well-designed geometric attention, which achieves improvements in all common metrics on MS-COCO dataset. By estimating the relevance between attention results and queries, Attention on Attention (AOA) module \cite{Huang2019AttentionOA} helps better model relationships between objects and filters out irrelative attention results. Pan et al. \cite{Pan2020XLinearAN} aim to improve the visual representations and promote multi-modal reasoning by modeling the $2^{nd}$ order interactions across multi-modal inputs. Global Enhanced Transformer (GET) \cite{Ji2021ImprovingIC} effectively supplements global context information, thereby generating high-quality captions. By integrating region features and grid features, more effective visual representations are learned by Dual-Level Collaborative Transformer (DLCT) \cite{Luo2021DualLevelCT} to generate more accurate caption sentences. Zhang et al. \cite{Zhang2021RSTNetCW} propose a relationship-sensitive transformer that includes a grid-augmented module and an adaptive-attention module. The grid-augmented module enhances visual representations by incorporating relative geometry features, and the adaptive-attention module adaptively guides the generation of the next word, thus more accurate image captioning descriptions are obtained. In Spatial and Scale-aware Transformer \cite{Zeng2022S2TF}, Zeng et al. propose a Spatial-aware Pseudo-supervised module to learn pseudo region features with a feature clustering manner, thus achieving more effective and interpretable image captioning. By taking the segmentation features as the complement information to enhance grid features, DIFNet \cite{WuDIFNetBV} generates caption sentences that are more faithful to given images and surpass the state-of-the-art captioning models. Moreover, pre-trained models \cite{Chen2021VisualGPTDA, Li2022BLIPBL} are also widely used to enhance image captioning.

Although current transformer-based captioning models \cite{Pan2020XLinearAN, Luo2021DualLevelCT, Qiu2021EgocentricIC, Wan2025FineGrainedIC} have achieved quite promising results, they are still prone to generating inaccurate and irrelevant descriptions due to the lack of contextual information and the over-reliance on generated partial descriptions for predicting the remaining words. Therefore, we introduce the segmentation feature to supplement the contextual information and then cooperate it with the region feature to enhance the visual and semantic representations for image captioning. Moreover, we design a novel brain-inspired dual-stream collaborative transformer to leverage these two features more effectively for more accurate and descriptive image captioning. 

\section{Preliminaries} 
Image captioning refers to generating a sentence to describe an image. In general, an image $I$ is described by a sentence  $Y$, which contains $T$ words denoted by  ${Y} = \left\{ {{y_1},{y_2}, \cdots ,{y_T}} \right\}$. Let $F$ denote the region feature extracted from image $I$, where $F$ consists of ${f_n}$ from $N$ regions, $n=1,...,N,$ and $ {f_n} \in {\mathbb{R} ^{{D_f}}}$. Current transformer-based image captioning models \cite{Pan2020XLinearAN, Luo2021DualLevelCT, WuDIFNetBV} first encoder the region feature $F$ and then use the encoded feature to guide the generation of next word $y_t$. These models process one word at each time step, which can be formulated as follows:

\begin{equation}
{y_t} = {D}\left( {{E}\left( F \right),{y_0},{y_1}, \cdots ,{y_{t - 1}}} \right)
\end{equation}where ${E}$ and $D$ denote the encoder and decoder, respectively. $E$ contains a sequence of multiple transformer layers, and each transformer layer consists of Multi-Head Self-Attention (MHSA) and Position-Wise Feed-Forward (PWFF) layers. The whole process of a transformer layer can be formulated as follows:

\begin{figure*}[t]
	\begin{center}
		\includegraphics[width=0.96\linewidth]{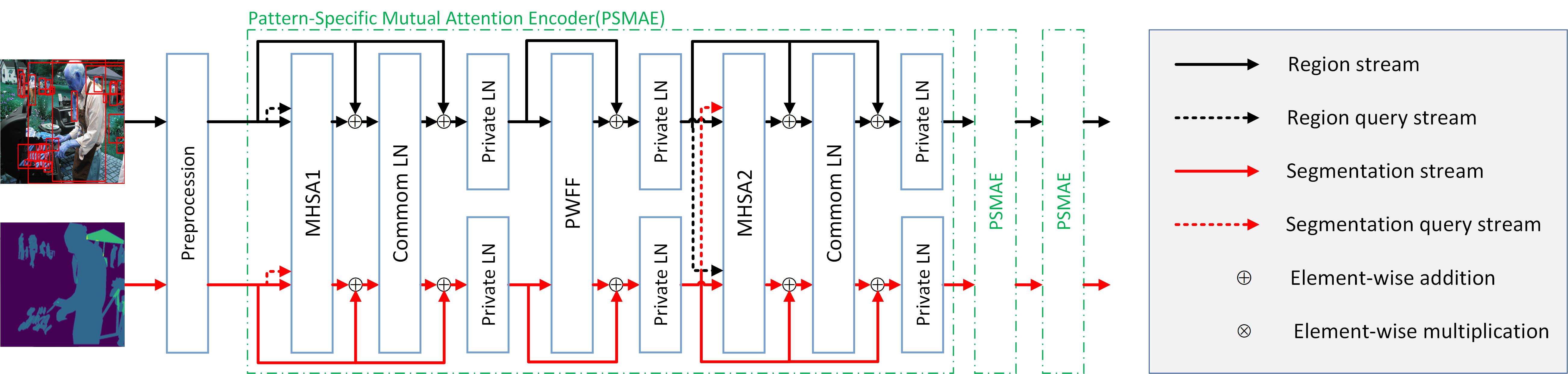}
	\end{center}
	\caption{The proposed Pattern-Specific Mutual Attention Encoder (PSMAE) consolidates each representation by highlighting its private information and querying the others to enhance the both representations for image captioning.}
	\label{encoder} 
\end{figure*}

\begin{equation}
\begin{array}{l}
	M^l = LN(LN({MHSA}(W_q^l{Z^l},W_k^l{Z^l},W_v^l{Z^l})\\
	{\kern 1pt} {\kern 1pt} {\kern 1pt} {\kern 1pt}{\kern 1pt} {\kern 1pt} {\kern 1pt} {\kern 1pt}{\kern 1pt} {\kern 1pt} {\kern 1pt} {\kern 1pt}{\kern 1pt} {\kern 1pt} {\kern 1pt} {\kern 1pt}{\kern 1pt} {\kern 1pt} {\kern 1pt} {\kern 1pt}{\kern 1pt} {\kern 1pt}  + {Z^l}) + {Z^l})
\end{array}
\end{equation}

\begin{equation}
{Z^{l + 1}} = LN\left( {PWFF\left( M^l \right) + M^l} \right)
\end{equation}where $Z$ is the sequence representation of the region feature $F$. $W_q^l$, $W_k^l$ and $W_v^l$ are embedding matrixes corresponding to \textbf{Query}, \textbf{Key} and \textbf{Values}, respectively. $LN$ denotes the Layer Normalization and $l$ denotes the $l$-th transformer layer of encoder $E$. We further use $\theta $ to denote the parameters of MHSA and PWFF, Eq. (2) and (3) can be reformulated as follows:

\begin{equation}
\begin{array}{l}
	M^l = LN(LN(MHSA({Z^l},{Z^l},{Z^l};\theta^l )\\
		{\kern 1pt} {\kern 1pt} {\kern 1pt} {\kern 1pt}{\kern 1pt} {\kern 1pt} {\kern 1pt} {\kern 1pt}{\kern 1pt} {\kern 1pt} {\kern 1pt} {\kern 1pt}{\kern 1pt} {\kern 1pt} {\kern 1pt} {\kern 1pt}{\kern 1pt} {\kern 1pt} {\kern 1pt} {\kern 1pt}{\kern 1pt} {\kern 1pt}+ {Z^l};{\alpha ^{l,0}},{\beta ^{l,0}}) + {Z^l};{\alpha ^{l,1}},{\beta ^{l,1}})
\end{array}
\end{equation}

\begin{equation}
{Z^{l + 1}} = LN(PWFF(M^l;\theta^l ) + M^l;{\alpha ^{l,2}},{\beta ^{{l,2}}})
\end{equation}where $\alpha$ and $\beta$ denote the scale and shift parameters of LN layer. By stacking multiple transformer layers, the enhanced visual feature (denoted by $H$) can be obtained and is ready for feeding into the language decoder to generate the output sequence.

The decoder $D$ is also composed of a sequence of multiple transformer layers, which includes both MHSA and PWFF layers. In the training stage, the decoder takes the output (denoted as $H$) of the encoder and the ground-truth word representations $T$ as inputs, and outputs the predicted word representations. To be specific, a Masked MHSA (denoted by $M\_MHSA$) layer is first used to capture the long-term dependence of words in the sentence, and the process can be formulated as follows:

\begin{equation}
\begin{array}{l}
{\hat T^k} = LN(LN(M\_MHSA({T^k},{T^k},{T^k};\theta^{l,0})\\
	{\kern 1pt} {\kern 1pt} {\kern 1pt} {\kern 1pt}{\kern 1pt} {\kern 1pt} {\kern 1pt} {\kern 1pt}{\kern 1pt} {\kern 1pt} {\kern 1pt} {\kern 1pt}{\kern 1pt} {\kern 1pt} {\kern 1pt} {\kern 1pt}{\kern 1pt} {\kern 1pt} {\kern 1pt} {\kern 1pt}{\kern 1pt} {\kern 1pt}+ {T^k};{\alpha ^{l,0}},{\beta ^{l,0}}) + {T^k};{\alpha ^{l,1}},{\beta ^{l,1}})
\end{array}
\end{equation}where $\theta^{l,0}_t$ denotes the parameters of the $l$-th MHSA layer. $\alpha ^{l,0}_t$, $\beta ^{l,0}_t$, $\alpha ^{l,1}_t$ and $\beta ^{l,1}_t$ are parameters of LN layers. Then, both $\hat T^k$ and enhanced visual features $H$ are used to generate the output sentence, which can be formulated as follows:

\begin{equation}
\begin{array}{l}
	{{\tilde T}^{k}} = LN(LN(MHA\left( {{{\hat T}^k},H,H;\theta^{l,1} } \right)\\
		{\kern 1pt} {\kern 1pt} {\kern 1pt} {\kern 1pt}{\kern 1pt} {\kern 1pt} {\kern 1pt} {\kern 1pt}{\kern 1pt} {\kern 1pt} {\kern 1pt} {\kern 1pt}{\kern 1pt} {\kern 1pt} + {Z^k};{\alpha ^{l,2}},{\beta ^{l,2}})+ {{\hat T}^k};{\alpha ^{l,3}},{\beta ^{l,3}}) + {{\hat T}^k};{\alpha ^{l,4}},{\beta ^{l,4}})
\end{array}
\end{equation}

\begin{equation}
{T^{k+1}} = LN\left( {PWFF\left( {{{\tilde T}^{k}};\theta^{l,1}} \right) + {{\tilde T}^{k}};{\alpha ^{l,5}},{\beta ^{l,5}}} \right)
\end{equation}where $MHA$ denotes the Multi-Head Attention. After that, the output of the last decoder ${\mathord{\buildrel{\lower3pt\hbox{$\scriptscriptstyle\smile$}} 
		\over T} }$ is mapped into the words through a linear operation and a softmax layer, which is expressed as follows:

\begin{equation}
Y = {{Softmax}}\left( {{{Linear}}\left( {\mathord{\buildrel{\lower3pt\hbox{$\scriptscriptstyle\smile$}} 
			\over T} } \right)} \right)
\end{equation}where $Y$ represents the generated sentence and ${\mathord{\buildrel{\lower3pt\hbox{$\scriptscriptstyle\smile$}} 
	\over T} }$ is the output of the last decoder. Finally, the loss between $Y$ and $Y^*$ is used to optimize the whole captioning model.

Our work exploits the segmentation feature as the semantic guidance to compensate the global information of the region feature, and the segmentation feature $S$ is extracted according to DIFNet \cite{WuDIFNetBV}. By introducing the segmentation feature, our paradigm can be formulated as follows:

 \begin{equation}
 	{y_t} = {DND}\left( {{PSMAE}\left( F, S \right),{y_0},{y_1}, \cdots ,{y_{t - 1}}} \right)
 \end{equation}where PSMAE and DND denote our proposed pattern-specific mutual attention encoder and dynamic nomination decoder, respectively, which are presented in the next section.
\section{Dual-Stream Collaborative Transformer}

The proposed DSCT contains multiple Pattern-Specific Mutual Attention Encoders (PSMAEs) and Dynamic Nomination Decoders (DNDs). Fig. \ref{dsct} gives the overview of our proposed DSCT, in which 3 PSMAEs and 3 DNDs are cascade connected. We begin with describing the PSMAE followed by DND, and then give out our objective functions.

\subsection{Pattern-Specific Mutual Attention Encoder}
Since the region feature provides object-level information and fine-grained descriptions to individual objects and the segmentation feature focuses more on global information and coarse-level context, both of them can be combined to enhance visual representations for image captioning. Inspired by the function of human brain that continuously consolidates and updates old memories, enabling it to search and use the most relevant memories to address new problems, we propose a Pattern-Specific Mutual Attention Encoder (PSMAE) to consolidate the both region and segmentation features before using them to generate caption sentences. The PSMAE is unified in a two-stream framework, in which information from each stream is used to guide the update of the other one. As shown in Fig. \ref{encoder}, the region and segmentation features are preprocessed before feeding into the PSMAE. PSMAE is designed to first highlight the private information of each representation (with the MHSA1 and the PWFF) and then update it by querying the other one (with the MHSA2). Thus, both region and segmentation features are consolidated by each other, and it is possible to generate a more accurate output sentence.

We follow \cite{Luo2021DualLevelCT, WuDIFNetBV} to extract the region feature and segmentation features, respectively. We denote the region input sequence as ${X_r} \in {^{N_r \times {d_{{\rm{model}}}}}}$ and the segmentation input sequence as ${X_s} \in {^{N_s \times {d_{{\rm{model}}}}}}$, where $N_r$ and $N_s$ are the numbers of elements of the corresponding feature. PSMAE contains two MHSA layers, a PWFF layer, and multiple private layer normalization (private LN) layers. As shown in Fig. \ref{encoder}, we first use a shared MHSA and a common LN layer to capture the common information across two streams, and then combine them with the original region and segmentation features, respectively. Thus, the newly obtained region or segmentation features contain both the private information of each stream and the common information across the two streams. Then, they are further highlighted by a shared PWFF layer and two private LN layers. The whole process can be formulated as follows:

\begin{equation}
\begin{array}{l}
	M_r^l = LN(LN(MHS{A^{l,0}}(Z_r^l,Z_r^l,Z_r^l;\theta _{rs}^{l})\\
	{\kern 1pt} {\kern 1pt} {\kern 1pt} {\kern 1pt}{\kern 1pt} {\kern 1pt} {\kern 1pt} {\kern 1pt}{\kern 1pt}{\kern 1pt} {\kern 1pt} {\kern 1pt} {\kern 1pt}{\kern 1pt} {\kern 1pt} {\kern 1pt} {\kern 1pt}{\kern 1pt}{\kern 1pt} {\kern 1pt} {\kern 1pt} {\kern 1pt}{\kern 1pt} {\kern 1pt} {\kern 1pt} {\kern 1pt}{\kern 1pt}
	+ Z_r^l;\alpha _r^{l,0},\beta _r^{l,0}) + Z_r^l;\alpha _r^{l,1},\beta _r^{l,1})
\end{array}
\end{equation}

\begin{equation}
\hat Z_r^{l + 1} = LN\left( {PWF{F^l}(M_r^l;\theta _{rs}^l) + M_r^l;\alpha _r^{l,2},\beta _r^{l,2}} \right)
\end{equation}

\begin{equation}
\begin{array}{l}
	M_s^l = LN(LN(MHS{A^{l,0}}(Z_s^l,Z_s^l,Z_s^l;\theta _{rs}^{l})\\
	{\kern 1pt} {\kern 1pt} {\kern 1pt} {\kern 1pt}{\kern 1pt} {\kern 1pt} {\kern 1pt} {\kern 1pt}{\kern 1pt}{\kern 1pt} {\kern 1pt} {\kern 1pt} {\kern 1pt}{\kern 1pt} {\kern 1pt} {\kern 1pt} {\kern 1pt}{\kern 1pt}{\kern 1pt} {\kern 1pt} {\kern 1pt} {\kern 1pt}{\kern 1pt} {\kern 1pt} {\kern 1pt} {\kern 1pt}{\kern 1pt}
	+ Z_s^l;\alpha _s^{l,0},\beta _s^{l,0}) + Z_s^l;\alpha _s^{l,1},\beta _s^{l,1})
\end{array}
\end{equation}

\begin{equation}
\hat Z_s^{l + 1} = LN\left( {PWF{F^l}(M_s^l;\theta _{rs}^l) + M_s^l;\alpha _s^{l,2},\beta _s^{l,2}} \right)
\end{equation}where $l$ varies from 0 to $L$  and $L$ denotes the number of PSMAE. $Z_r^0$ and $Z_s^0$ equal to $X_r$ and $X_s$, respectively. $\theta _{rs}^l$ denotes the model parameter of ${MHSA}^{l,0}$, ${MHSA}^{l,1}$ and ${PWFF}^l$, $\alpha$ and  $\beta $ are the corresponding scale and shift parameters of private LN layers. 0, 1 and 2 denote the first, second and third private LN layer, respectively. After that, PSMAE further updates these two features by querying the other one and the process can be formulated as follows:

\begin{figure*}[t]
	\begin{center}
		\includegraphics[width=0.96\linewidth]{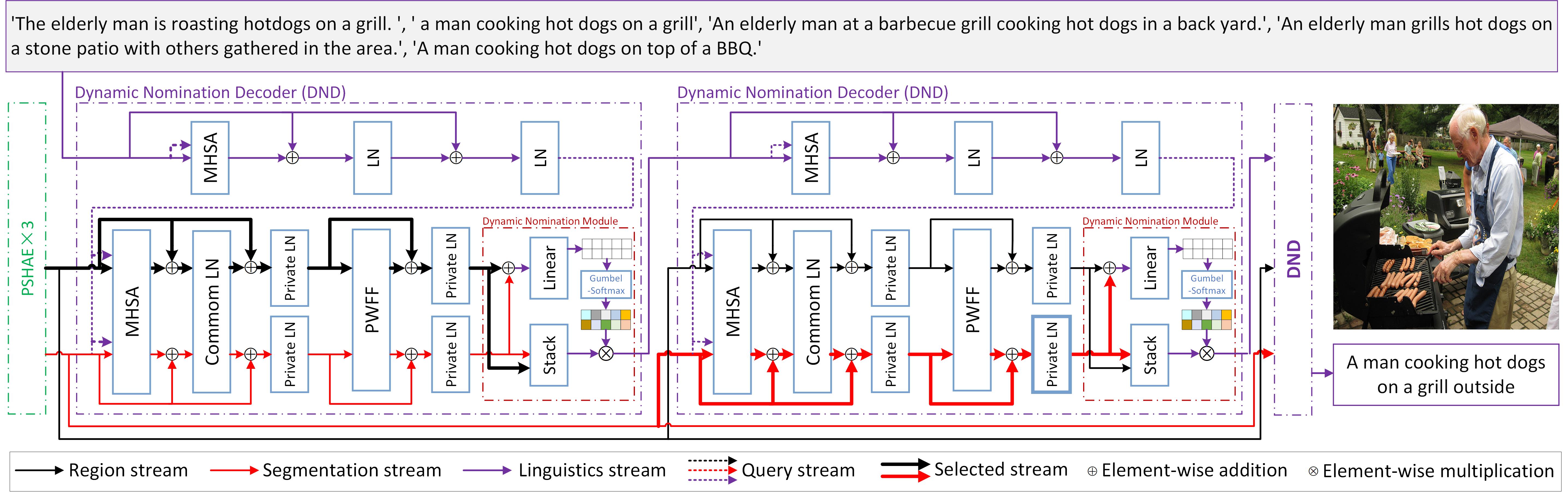}
	\end{center}
	\caption{The proposed Dynamic Nomination Decoder(DND) is a two-stream structure in which an proper stream is dynamically nominated to generate the next words by integrating a Dynamic Nomination Module. This design forces each stream to update its private information to learn more effective visual and semantic representations. Moreover, by stacking multiple DNDs, the semantic inconsistencies and spatial misalignment issues between region and segmentation features are bypassed, so that they are utilized more effectively for improving image captioning performance. }
	\label{decoder}
\end{figure*}

\begin{equation}
\begin{array}{l}
	Z_r^{l+1} = LN(LN(MH{A^{l,1}}(\hat Z_s^{l + 1},\hat Z_r^{l + 1},\hat Z_r^{l + 1};\theta _{rs}^l)\\
	{\kern 1pt} {\kern 1pt} {\kern 1pt} {\kern 1pt}{\kern 1pt} {\kern 1pt} {\kern 1pt} {\kern 1pt}{\kern 1pt}{\kern 1pt} {\kern 1pt} {\kern 1pt} {\kern 1pt}{\kern 1pt} {\kern 1pt} {\kern 1pt} {\kern 1pt}{\kern 1pt}{\kern 1pt} {\kern 1pt} {\kern 1pt}
	+ \hat Z_r^{l + 1};\alpha _r^{l,3},\beta _r^{l,3}) + \hat Z_r^{l + 1};\alpha _r^{l,4},\beta _r^{l,4})
\end{array}
\end{equation}

\begin{equation}
\begin{array}{l}
	Z_s^{l+1} = LN(LN(MH{A^{l,1}}(\hat Z_r^{l + 1},\hat Z_s^{l + 1},\hat Z_s^{l + 1};\theta _{rs}^l)\\
	{\kern 1pt} {\kern 1pt} {\kern 1pt} {\kern 1pt}{\kern 1pt} {\kern 1pt} {\kern 1pt} {\kern 1pt}{\kern 1pt}{\kern 1pt} {\kern 1pt} {\kern 1pt} {\kern 1pt}{\kern 1pt} {\kern 1pt} {\kern 1pt} {\kern 1pt}{\kern 1pt}{\kern 1pt} {\kern 1pt} {\kern 1pt}
	+ \hat Z_s^{l + 1};\alpha _s^{l,3},\beta _s^{l,3}) + \hat Z_s^{l + 1};\alpha _s^{l,4},\beta _s^{l,4})
\end{array}
\end{equation}

In this way, the region feature is updated by referring to the segmentation feature, and then the segmentation feature is boosted according to the updated region feature. Therefore, the region feature learns more effective contextual information from the segmentation feature, which in turn helps to update the segmentation feature. Likewise, we use a common LN layer to capture the common information across the two streams, which are fused by the element-wise addition operation. Then, two private LN layers are further used to highlight their private information. Hence, with the proposed PSMAE, both the region feature and segmentation feature are boosted for image captioning.

\subsection{Dynamic Nomination Decoder}
With the proposed PSMAE, we obtain the consolidated region feature ${\mathord{\buildrel{\lower3pt\hbox{$\scriptscriptstyle\frown$}} 
		\over Z} _r}$  and segmentation feature ${\mathord{\buildrel{\lower3pt\hbox{$\scriptscriptstyle\frown$}} 
		\over Z} _s}$. Next, we search the most relevant memories from the old knowledge (i.e., the consolidated region and segmentation features) according to the partially generated sentences, and then use them to guide the generation of caption sentences. As the region and segmentation features are actually pattern-specific features, due to their semantic inconsistencies and spatial misalignment, it is difficult to directly fuse/use them for image captioning. Moreover, current captioning models \cite{Pan2020XLinearAN, Luo2021DualLevelCT, WuDIFNetBV} depend highly on the partially generated words for predicting the remaining words, which may lead to irrelevant descriptions. Therefore, this paper proposes a Dynamic Nomination Decoder (DND) to cooperate with our PSMAE to alleviate this problem. The DND is also a two-stream structure, including a region feature stream and a segmentation feature stream. With a well-designed Dynamic Nomination Module (DNM), the proposed DND dynamically selects the most relevant learning blocks and exploits the homogeneous features between the region and segmentation feature streams for achieving more accurate and descriptive image captioning.

In the training stage, the input of DND includes the output of PSMAE (i.e., region feature ${\mathord{\buildrel{\lower3pt\hbox{$\scriptscriptstyle\frown$}} 
		\over Z} _r}$  and segmentation feature ${\mathord{\buildrel{\lower3pt\hbox{$\scriptscriptstyle\frown$}} 
		\over Z} _s}$) and the ground-truth word representations (denoted as ${Z_t}$, ${Z_t} \in {R^{{\rm{seq}} \times {d_{{\rm{model}}}}}}$,  where $seq$ denotes the length of sentence). The word representation is firstly added with the positional embedding and then a $M\_MHSA$ is performed, the process is formulated as follows:

\begin{equation}
\begin{array}{l}
	\hat Z_t^l = LN(LN(M\_MHSA^{l}_t(Z_t^l,Z_t^l,Z_t^l;\theta _t^l)\\
	{\kern 1pt} {\kern 1pt} {\kern 1pt} {\kern 1pt}{\kern 1pt} {\kern 1pt} {\kern 1pt} {\kern 1pt}{\kern 1pt}{\kern 1pt} {\kern 1pt} {\kern 1pt} {\kern 1pt}{\kern 1pt} {\kern 1pt} {\kern 1pt} {\kern 1pt}{\kern 1pt}{\kern 1pt} {\kern 1pt} {\kern 1pt}
	+ Z_t^l;\alpha _t^{l,0},\beta _t^{l,0}) + Z_t^l;\alpha _t^{l,1},\beta _t^{l,1})
\end{array}
\end{equation}where $\theta $, $\alpha$ and  $\beta $ are the corresponding parameters. Then, both the region and segmentation features are leveraged to guide the generation of image caption sentences. As the generation of the next token should only depend on the previously generated partial sentences, the masked self-attention operation \cite{Cornia2020MeshedMemoryTF, WuDIFNetBV} is adopted. Similar to the design of PSMAE, DND also uses a shared MHA and PWFF layers and private LN layer to handle the information of the two streams as shown in Fig. \ref{decoder}. The whole process can be formulated as follows:
 
\begin{equation}
\begin{array}{l}
	M_{rt}^l = LN(LN(MHA_{rs}^l(\hat Z_t^l,{{\mathord{\buildrel{\lower3pt\hbox{$\scriptscriptstyle\frown$}} 
				\over Z} }_r},{{\mathord{\buildrel{\lower3pt\hbox{$\scriptscriptstyle\frown$}} 
				\over Z} }_r};\theta _{rs}^l)\\
			{\kern 1pt} {\kern 1pt} {\kern 1pt} {\kern 1pt}{\kern 1pt} {\kern 1pt} {\kern 1pt} {\kern 1pt}{\kern 1pt}{\kern 1pt} {\kern 1pt} {\kern 1pt} {\kern 1pt}{\kern 1pt} {\kern 1pt} {\kern 1pt} {\kern 1pt}{\kern 1pt}{\kern 1pt} {\kern 1pt} {\kern 1pt}
	+ \hat Z_t^l;\alpha _{rs}^{l,0},\beta _{rs}^{l,0}) + \hat Z_t^l;\alpha _r^{l,0},\beta _r^{l,0})
\end{array}
\end{equation}	

\begin{equation}
	{Z_{tr}^l} = LN\left( {PWF{F^l}(M_{rt}^l;\theta _{rs}^{l}) + M_{rt}^l;\alpha _{r}^{l,1},\beta _{r}^{l,1}} \right)
\end{equation}

\begin{equation}
	\begin{array}{l}
		M_{st}^l = LN(LN(MHA_{rs}^l(\hat Z_t^l,{{\mathord{\buildrel{\lower3pt\hbox{$\scriptscriptstyle\frown$}} 
					\over Z} }_s},{{\mathord{\buildrel{\lower3pt\hbox{$\scriptscriptstyle\frown$}} 
					\over Z} }_s};\theta _{rs}^l)\\
		{\kern 1pt} {\kern 1pt} {\kern 1pt} {\kern 1pt}{\kern 1pt} {\kern 1pt} {\kern 1pt} {\kern 1pt}{\kern 1pt}{\kern 1pt} {\kern 1pt} {\kern 1pt} {\kern 1pt}{\kern 1pt} {\kern 1pt} {\kern 1pt} {\kern 1pt}{\kern 1pt}{\kern 1pt} {\kern 1pt} {\kern 1pt}
		+ \hat Z_t^l;\alpha _{rs}^{l,0},\beta _{rs}^{l,0}) + \hat Z_t^l;\alpha _s^{l,0},\beta _s^{l,0})
	\end{array}
\end{equation}

\begin{equation}
	Z_{ts}^l = LN\left( {PWF{F^l}(M_{st}^l;\theta _{rs}^l) + M_{st}^l;\alpha _{s}^{l,1},\beta _{s}^{l,1}} \right)
\end{equation}where $\theta $ represents the parameters of shared MHA and PWFF layers, and $\alpha$ and  $\beta $ are the corresponding parameters of LN layers.

\textbf{Dynamic Nomination Module. } The outputs of the two streams (i.e., $Z_{tr}^l$ and $Z_{ts}^l$) are inputted into a Dynamic Nomination Module (DNM). The aim of DNM is to select the most suitable features from these two streams to generate the next word. DNM first fuses the candidate features $\left\{ {Z_{tr}^l,Z_{ts}^l} \right\}$ with an element-wise addition operation and then adopts a linear layer to obtain a tensor ${\Gamma }$ of shape  ${\rm{seq}} \times 2$. The last dimension of ${\Gamma }$ represents the probability of nominating features of each type from two candidates. After that, DNM uses element-wise hard sampling on ${\Gamma }$ to calculate the nomination map (denoted as $\Psi$). The process of obtaining the nominated feature index for the $i$-th word can be formulated as follows:

\begin{equation}
{w_i} = \arg \max \left( {{\tau _{i,1}},{\tau _{i,2}}} \right)
\end{equation}where ${\tau _{i,c}}$ denotes the value of ${\Gamma }$ corresponding to indexes $(i, c)$. With the argmax function, the nomination map $\Psi$ can be obtained. That means, for each word in the sentence, the DNM nominates the candidate features by setting the selected one to 1 and the other one to 0. For example, if the region feature is nominated for the $i$-th word, the corresponding nomination vector ${\Psi _i}$ is  $[1,0]$. After that, we follow \cite{Liu2022NomMerNS} and use the Gumbel Softmax trick \cite{Jang2017CategoricalRW} to address the non-differentiable problem of the hard sampling process. Finally, the updated word representations are calculated by using nomination map $\Psi$ by:

\begin{equation}
Z_t^{l + 1} = Z_{tr}^l{\Psi _{,0}} + Z_{ts}^l{\Psi _{,1}}
\end{equation}where ${\Psi _{,0}}$ denotes the nominating results of all words corresponding to region features, and  ${\Psi _{,1}}$ denotes the nominating results corresponding to segmentation features. 

In each DND, the most appropriate stream is nominated to guide the generation of the next token. Therefore, DSCT is able to fuse region and segmentation features by stacking multiple DNDs, i.e., the best pathway can be dynamically nominated by our DSCT as shown in Fig. \ref{decoder}. Compared with using element-wise addition or concatenation operation to fuse them, the proposed DND effectively bypasses their semantic inconsistencies and spatial misalignment issues. On the other hand, by introducing the dynamic nomination module, the DND also forces each stream to learn more private information, which further helps to improve the visual representation for image captioning.

\subsection{Dual-Stream Collaborative Transformer}
The proposed PSMAE consolidates and updates both region feature and segmentation feature by highlighting their private information and querying each other. Then, the DND is designed to fuse the consolidated region and segmentation features in a dynamic way such that the homogeneous features between them are mined to enhance the visual representation and reduce the dependence on the generated partial descriptions for predicting the remaining words. By integrating the PSMAE and DND into a Dual-Stream Collaborative Transformer (DSCT), as shown in Fig. \ref{dsct}, via a seamless formulation, the segmentation and region features are fully utilized to improve both representations. This leads to our proposed DSCT outperforming the state-of-the-art image captioning models. 

\begin{table*}
	\caption{Performance comparisons with the state-of-the-art image captioning models on COCO Karpathy test split, where B@N, M, R, C and S are short for BLEU@N, METEOR, ROUGE-L, CIDEr and SPICE scores. All values are reported as percentage (\%). $\Sigma $ indicates model ensemble/fusion.}
	\begin{center}
		\footnotesize
		\begin{tabular}{p{3.8cm}|p{0.4cm}p{0.4cm}p{0.4cm}p{0.4cm}p{0.4cm}p{0.4cm}p{0.4cm}p{0.4cm}|p{0.4cm}p{0.4cm}p{0.4cm}p{0.4cm}p{0.4cm}p{0.4cm}p{0.4cm}p{0.5cm}}
			\hline
			\multirow{2}{*}{Method} & \multicolumn{8}{|c}{Cross-Entropy Loss} & \multicolumn{8}{|c}{CIDEr Score Optimization}  \\
			\cline{2-17}
			&B@1 &B@2  &B@3 &B@4 &M &R &C &S  &B@1 &B@2  &B@3 &B@4 &M &R &C &S\\
			\cline{2-17}	
			\cline{0-0}
			LSTM{$\rm _{CVPR15}$} \cite{Vinyals2015ShowAT} 		& - & - & - & 29.6 &25.2 & 52.6 &94.0 & - & - & - & - & 31.9 &25.5 & 54.3 &106.3 & -\\	
			SCST{$\rm _{CVPR17}$} \cite{Rennie2017SelfCriticalST} 		& - & - & - &30.0 &25.9 &53.4 &99.4 & - & - & - & - &34.2 &26.7 &55.7 &114.0 & -\\
			LSTM-A{$\rm _{ICCV17}$} \cite{Yao2017BoostingIC} 		&75.4 & - & - &35.2 &26.9 &55.8 &108.8 &20.0 &78.6 & - & - &35.5 &27.3 &56.8 &118.3 &20.8\\
			RFNet{$\rm _{ECCV18}$} \cite{Jiang2018RecurrentFN} 		&76.4 &60.4 &46.6 &35.8 &27.4 &56.5 &112.5 &20.5 &79.1 &63.1 &48.4 &36.5 &27.7 &57.3 &121.9 &21.2\\
			Up-Down{$\rm _{CVPR18}$} \cite{Anderson2018BottomUpAT} 	& 77.2 & - & - &36.2 &27.0 &56.4 &113.5 &20.3 & 79.8 & - & - &36.3 &27.7 &56.9 &120.1 &21.4\\
			GCN-LSTM{$\rm _{ECCV18}$} \cite{Yao2018ExploringVR} 	& 77.3 & - & - &36.8 &27.9 &57.0 &116.3 &20.9 &80.5 & - & - &38.2 &28.5 &58.3 &127.6 &22.0\\
			LBPF{$\rm _{CVPR19}$} \cite{Qin2019LookBA} 		& 77.8 & - & - &37.4 &28.1 &57.5 &116.4 &21.2 &80.5 & - & - &38.3 &28.5 &58.4 &127.6 &22.0\\
			SGAE{$\rm _{CVPR18}$} \cite{Yang2019AutoEncodingSG} 		& 77.6 & - & - &36.9 &27.7 &57.2 &116.7 &20.9 & - & - & - &38.4 &28.4 &58.6 &127.8 &22.1\\
			AoANet{$\rm _{ICCV19}$} \cite{Huang2019AttentionOA} 		& 77.4 & - & - &37.2 &28.4 &57.5 &119.8 &21.3 &80.2 & - & - &38.9 &29.2 &58.8 &129.8 &22.4\\ 
			M2 Transformer{$\rm _{CVPR20}$} \cite{Cornia2020MeshedMemoryTF} & - & - & - &- &- &- &- &- &80.8 & - & - &39.1 &29.2 &58.6 &131.2 &22.6\\ 
			X-Transformer{$\rm _{CVPR20}$} \cite{Pan2020XLinearAN} &77.3 &61.5 &47.8 &37.0 &28.7 &57.5 &120.0 &21.8  &80.9 & 65.8 & 51.5 &39.7 &29.5 &59.1 &132.8 &23.4\\ 
			DLCT{$\rm _{AAAI21}$} \cite{Luo2021DualLevelCT} & - & - & - &- &- &- &- &-  &81.4 & - & -&39.8 &29.5 &59.1 &133.8 &23.0\\ 
			RSTNet{$\rm _{CVPR21}$} \cite{Zhang2021RSTNetCW} & - & - & - &- &- &- &- &-  &81.1 & - & - &39.3 &29.4 &58.8 &133.3 &23.0\\ 
			DIFNet{$\rm _{CVPR22}$} \cite{WuDIFNetBV} & - & - & - &- &- &- &- &-  &81.7 & - & - &40.0 &29.7 &59.4 &136.2 &23.2\\ \hline
			\textbf{Our DSCT} & 80.5 & - & - &37.9 &29.7 &58.9 &123.6 &22.3  &\textbf{82.7} & - & - &\textbf{40.3} &\textbf{30.5} &\textbf{59.9} &\textbf{137.6} &\textbf{23.9}\\ \hline
			&Ensemble/Fusion\\ \hline
			$\rm SCST^\Sigma ${$\rm _{CVPR17}$} \cite{Rennie2017SelfCriticalST}		& - & - & - &32.8 &26.7 &55.1 &106.5 & - & - & - & - &35.4 &27.1 &56.6 &117.5 & -\\
			$\rm RFNet^\Sigma ${$\rm _{ECCV18}$} \cite{Jiang2018RecurrentFN}		&77.4 &61.6 &47.9 &37.0 &27.9 &57.3 &116.3 &20.8 &80.4 &64.7 &50.0 &37.9 &28.3 &58.3 &125.7 &21.7\\
			$\rm GCN-LSTM^\Sigma ${$\rm _{ECCV18}$} \cite{Yao2018ExploringVR} 	&77.4 &- &- &37.1 &28.1 &57.2 &117.1 &21.1 &80.9 &- &- &38.3 &28.6 &58.5 &128.7 &22.1\\
			$\rm SGAE^\Sigma ${$\rm _{CVPR18}$} \cite{Yang2019AutoEncodingSG} 	&- &- &- &- &- &- &- &- &81.0 &- &- &39.0 &28.4 &58.9 &129.1 &22.2\\
			$\rm HIP^\Sigma ${$\rm _{ICCV19}$} \cite{Yao2019HierarchyPF} 	&- &- &- &38.0 &28.6 &57.8 &120.3 &21.4 &- &- &- &39.1 &28.9 &59.2 &130.6 &22.3\\
			$\rm AoANet^\Sigma ${$\rm _{ICCV19}$} \cite{Huang2019AttentionOA} 	&78.7 &- &- &38.1 &28.5 &58.2 &122.7 &21.7 &81.6 &- &- &40.2 &29.3 &59.4 &132.0 &22.8\\ 
			$\rm X-Transformer^\Sigma ${$\rm _{CVPR20}$} \cite{Pan2020XLinearAN}	&77.8 &62.1 &48.6 &37.7 &29.0 &58.0 &122.1 &21.9 &81.7 &66.8 &52.6 &40.7 &29.9 &59.7 &135.3 &23.8\\
			$\rm DLCT^\Sigma ${$\rm _{AAAI21}$} \cite{Luo2021DualLevelCT} & - & - & - &- &- &- &- &-  &82.2 &- &- &40.8 &29.9 &59.8 &137.5 &23.3\\ \hline
			\textbf{Our DSCT} & 80.9 & - & - &38.5 &30.1 &59.2 &124.7 &22.6  &\textbf{83.1} &- &- &\textbf{41.6} &\textbf{30.4} &\textbf{60.5} &\textbf{139.3} &\textbf{24.2}\\ \hline
		\end{tabular}
	\end{center}
	\label{taboff}
\end{table*}

\subsection{Objective Functions}
Following \cite{Anderson2018BottomUpAT, Ranzato2016SequenceLT, Rennie2017SelfCriticalST, WuDIFNetBV}, we first use the cross-entropy loss function to pre-train our DSCT, and then use the CIDEr-D score \cite{Anderson2018BottomUpAT} as the reward of reinforcement learning to finetune its sequence generation process. The process of the pre-training stage can be formulated as follows:

\begin{equation}
{L_{XE}} =  - \sum\limits_{t = 1}^T {\log \left( {{p_\theta }\left( {y_t^ * \left| {y_{1:t - 1}^ * } \right.} \right)} \right)} 
\end{equation} where $y_{1:T}^ * $ denote the ground-truth caption sentence, and $\theta$ is the parameters of our DSCT. Then, the pre-trained DSCT is further finetuned with the CIDEr-D score, which can be expressed as follows:

\begin{equation}
{\nabla _\theta }{L_{RL}}\left( \theta  \right) =  - \frac{1}{k}\sum\limits_{i = 1}^k {\left( {r\left( {y_{1:T}^i - b} \right)} \right)} {\nabla _\theta }{\log _{{p_\theta }}}\left( {y_{1:T}^i} \right)
\end{equation}

\begin{equation}
	b = \sum\nolimits_i {r\left( {y_{1:T}^i} \right)/k}
\end{equation}where $k$ is the beam size, $r$ is the CIDEr-D function and $b$ denotes the mean of the rewards.

\section{Experiments}
In this section, the datasets and implementation details are first introduced. Then, we compare our proposed DSCT with the state-of-the-art image captioning models \cite{Huang2019AttentionOA, Cornia2020MeshedMemoryTF, Pan2020XLinearAN, Luo2021DualLevelCT, Zhang2021RSTNetCW}. After that, ablation studies and self-evaluations are conducted. Finally, we present the experimental results and discussions.

\subsection{Datasets and Implementation details}
\textbf{Datasets.} Our DSCT is evaluated on image captioning benchmark dataset MS-COCO \cite{Lin2014MicrosoftCC}. It contains 164,062 images in total, and each image is annotated with 5 captions. MS-COCO dataset is then divided into three subsets, i.e., training set (82,783 images), validation set (40,504 images) and testing set (40,775 images). We evaluate our DSCT in both online and offline ways. The online one is done through MS COCO online testing server as the annotations are not provided. For the offline evaluation, we use the Karpathy splits \cite{Karpathy2017DeepVA} that include 5000 validation images, 5000 testing images, and the remaining images for training. 

\textbf{Evaluation Metrics.} The popular image captioning metrics includes BLEU-N(B@N) \cite{Papineni2002BleuAM}, METEOR(M) \cite{Lavie2007METEORAA}, ROUGE(R) \cite{Lavie2007METEORAA}, CIDEr(C) \cite{Vedantam2015CIDErCI}, and SPICE(S) \cite{Anderson2016SPICESP}, and we follow the standard evaluation protocol and use the above all captioning metrics.

\textbf{Implementation Details.} We follow \cite{Cornia2020MeshedMemoryTF} and \cite{WuDIFNetBV} to separately extract the region feature and segmentation feature, which are both denoted by 2048-dimensional feature vectors. We use the one-hot vector to denote the word representation and the sinusoidal positional encoding to represent the word position, they are summed and then fed into the first DND.

In the implementation of DSCT, the $d_{model}$ is set to 512, and the number of heads and the batch size are equal to 8 and 50, respectively. Our DSCT contains 3 PSMAEs and 3 DNDs, and the beam size is set to 5. The dropout operation is also used in our DSCT and the keep probability is set to 0.9. We follow \cite{WuDIFNetBV} to set the learning rate scheduling strategy and Adam optimizer is selected. Our DSCT is trained with Pytorch on an Nvidia Tesla V100 GPU. We show the comparisons with state-of-the-art captioning methods below.

\begin{table*}
	\caption{Leaderboard of the published state-of-the-art image captioning models on the COCO online testing server, where B@N, M, R and C are short for BLEU@N, METEOR, ROUGE-L and CIDEr scores. All values are reported as percentage (\%).}
	\begin{center}
		\footnotesize
		\begin{tabular}{p{3.5cm}|p{0.5cm}p{0.5cm}p{0.5cm}p{0.5cm}p{0.5cm}p{0.5cm}p{0.5cm}p{0.5cm}|p{0.5cm}p{0.5cm}p{0.5cm}p{0.5cm}p{0.5cm}p{0.5cm}}
			\hline
			\multirow{2}{*}{Method} & \multicolumn{2}{c}{B@1} & \multicolumn{2}{c}{B@2} & \multicolumn{2}{c}{B@3} & \multicolumn{2}{c}{B@4} &\multicolumn{2}{c}{M} & \multicolumn{2}{c}{R} &\multicolumn{2}{c}{C} \\			\cline{2-15}
			&c5 &c40 &c5 &c40 &c5 &c40 &c5 &c40 &c5 &c40 &c5 &c40 &c5 &c40\\
			\cline{2-15}	
			\cline{0-0}
			SCST{$\rm _{CVPR17}$} \cite{Rennie2017SelfCriticalST} &78.1 &93.7 &61.9 &86.0 &47.9 &75.9 &35.2 &64.5 &27.0 &35.5 &56.3 &70.7 &114.7 &116.7 \\
			Up-Down{$\rm _{CVPR18}$} \cite{Anderson2018BottomUpAT}  &80.2 &95.2 &64.1 &88.8 &49.1 &79.4 &36.9 &68.5 &27.6 &36.7 &57.1 &72.4 &117.9 &120.5 \\
			RFNet{$\rm _{ECCV18}$} \cite{Jiang2018RecurrentFN}  &80.4 &95.0 &64.9 &89.3 &50.1 &80.1 &38.0 &69.2 &28.2 &37.2 &58.2 &37.1 &122.9 &125.1 \\
			GCN-LSTM{$\rm _{ECCV18}$} \cite{Yao2018ExploringVR} &80.8 &95.9 &65.5 &89.3 &50.8 &80.3 &38.7 &69.7 &28.5 &37.6 &58.5 &73.4 &125.3 &126.5\\
			SGAE{$\rm _{CVPR18}$} \cite{Yang2019AutoEncodingSG}  &81.0 &95.3 &65.6 &89.5 &50.7 &80.4 &38.5 &69.7 &28.2 &37.2 &58.6 &73.6 &123.8 &126.5\\
			ETA{$\rm _{ICCV19}$} \cite{Li2019EntangledTF}  &81.2 &95.0 &65.5 &89.0 &50.9 &80.4 &38.9 &70.2 &28.6 &38.0 &58.6 &73.9 &122.1 &124.4\\
			AoANet{$\rm _{ICCV18}$} \cite{Huang2019AttentionOA}  &81.0 &95.0 &65.8 &89.6 &51.4 &81.3 &39.4 &71.2 &29.1 &38.5 &58.9 &74.5 &126.9 &129.6\\
			M2 Transformer{$\rm _{CVPR20}$} \cite{Cornia2020MeshedMemoryTF}  &81.6 &96.0 &66.4 &90.8 &51.8 &82.7 &39.7 &72.8 &29.4 &39.0 &59.2 &74.8 &129.3 &132.1\\
			X-Transformer{$\rm _{CVPR20}$} \cite{Pan2020XLinearAN} &81.3 &95.4 &66.3 &90.0 &51.9 &81.7 &39.9 &71.8 &29.5 &39.0 &59.3 &74.9 &129.3 &131.4\\ 
			DLCT{$\rm _{AAAI21}$} \cite{Luo2021DualLevelCT} &82.0 &96.2 &66.9 &91.0 &52.3 &83.0 &40.2 &73.2 &29.5 &39.1 &59.4 &74.8 &131.0 &133.4\\
			RSTNet{$\rm _{CVPR21}$} \cite{Zhang2021RSTNetCW} &81.7 &96.2 &66.5 &90.9 &51.8 &82.7 &39.7 &72.5 &29.3 &38.7 &59.2 &74.2 &130.1 &132.4\\ 
			$S^2$-Transformer{$\rm _{IJCAI22}$} \cite{Zeng2022S2TF} &81.9 &96.4 &66.7 &91.3 &52.1 &83.1 &40.0 &73.1 &29.5 &39.2 &59.2 &74.7 &131.5 &134.5\\ \hline
			\textbf{Our DSCT} &\textbf{82.9} &\textbf{97.3} &\textbf{67.7} &\textbf{91.5} &\textbf{53.1} &\textbf{83.8} &\textbf{40.6} &\textbf{73.9}  &\textbf{30.2} &\textbf{39.8} &\textbf{59.9}&\textbf{75.3} &\textbf{133.1} &\textbf{136.7} \\ \hline
		\end{tabular}
	\end{center}
	\label{tabon}
\end{table*}

\subsection{Performance Comparison}
\textbf{Offline Evaluation. } The performance comparison results on the offline COCO Karpathy test split with the state-of-the-art captioning models \cite{Huang2019AttentionOA, Cornia2020MeshedMemoryTF, Pan2020XLinearAN, Luo2021DualLevelCT, Zhang2021RSTNetCW, WuDIFNetBV} are shown in Tabel \ref{taboff}. As the training of image captioning models contains two stages, for fair comparisons, we separately report the experimental results for each stage. Meanwhile, following \cite{Pan2020XLinearAN, Luo2021DualLevelCT}, we also present the performance of single model and ensemble/fused models. As shown in Table \ref{taboff}, we can find that our proposed DSCT outperforms the state-of-the-art captioning methods, which includes region feature-based models \cite{Anderson2018BottomUpAT, Huang2019AttentionOA, Cornia2020MeshedMemoryTF, Pan2020XLinearAN}, grid feature-based models \cite{Jiang2020InDO, Luo2021DualLevelCT, WuDIFNetBV} and fusion feature-based models \cite{Luo2021DualLevelCT, WuDIFNetBV}. For single models, our proposed DSCT achieves 137.6\% CIDEr score, which surpasses the current best captioning model-DIFNet. When using ensemble models, our DSCT even achieves 139.3\% CIDEr score that beats the state-of-the-art captioning models \cite{ Yao2019HierarchyPF, Huang2019AttentionOA, Pan2020XLinearAN, Luo2021DualLevelCT}. Moreover, for other evaluation metrics including BLEU-N(B@N), METEOR(M), ROUGER and SPICE(S), our proposed DSCT also obtains the best scores as shown in Tabel \ref{taboff}. These performance improvements indicate that 1) the proposed PSMAE highlights the private information of each representation, and both region features and segmentation features are updated to improve image captioning performance. 2) With the well-designed DND, the region and segmentation representations are dynamically nominated to generate the next word, at the same time, this design further forces different streams to focus on their private information, which in turn boosts both streams. 3) By integrating the PSMAE and DND via the proposed dual-stream collaborative transformer, more effective visual and semantic representations are learned for more accurate and descriptive image captioning.

\textbf{Online Evaluation. } The captioning performance of our proposed DSCT on the online COCO test server is reported below. Following \cite{Cornia2020MeshedMemoryTF, Pan2020XLinearAN, Luo2021DualLevelCT}, we first train 4 DSCT models separately on the Karpathy training split \cite{Karpathy2017DeepVA}, and then submit them to the online COCO test server. We give both comparisons with 5 reference captions (c5) and 40 reference captions, and the corresponding experimental results are shown in Table \ref{tabon}. From Table \ref{tabon}, we can find that our ensemble DSCT outperforms the state-of-the-art captioning models \cite{Huang2019AttentionOA, Cornia2020MeshedMemoryTF, Pan2020XLinearAN, Luo2021DualLevelCT, Zhang2021RSTNetCW}. This indicates that our DSCT generates more accurate and descriptive captions by utilizing the region feature and segmentation feature more effectively in a novel dynamic way via the proposed dual-stream collaborative transformer.

\begin{figure*}[!t]
	\centering
	\includegraphics[width=0.98\linewidth]{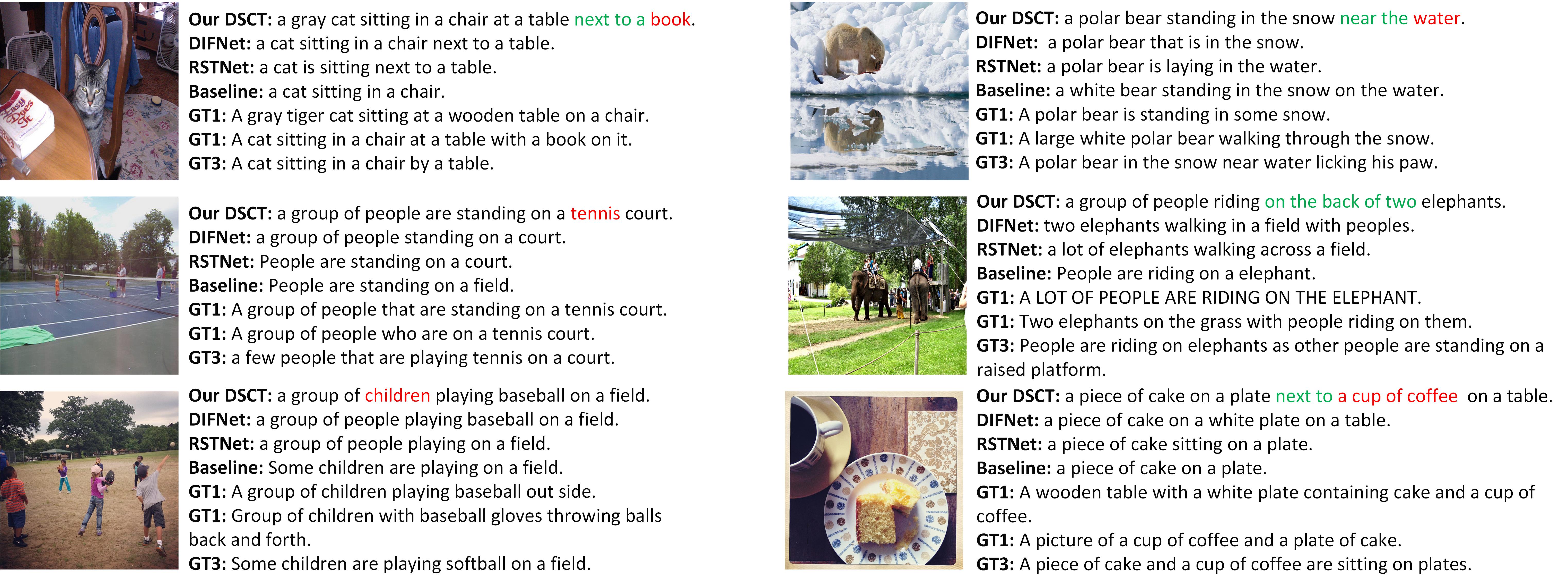}
	\centering
	\caption{Examples of image captioning results by baselines (such as DIFNet, RSTNet and our baseline transformer) and our proposed DSCT, along with the ground-truth sentences. Compared to other baselines, our proposed DSCT can identify more objects (red words) and accurately describe the relationships between objects (green words) in images, thereby generating more accurate and decriptive sentences. }
	\label{example}
\end{figure*}

\textbf{Qualitative Analysis.} We show several image captioning results of our baseline transformer and our DSCT in Figure. \ref{example}. GT1, GT2 and GT3 are the ground-truth captions. The baseline and our DSCT use the same region features. Compared with the baseline, our proposed DSCT achieves more accurate and descriptive image captioning by leveraging the region feature and the segmentation feature in a novel dynamic way. As shown in the first column of Figure. \ref{example}, we can see that our proposed DSCT gives more accurate and comprehensive descriptions about the visual objects in the image. For example, the baseline generates the caption sentence ``some children are playing on a field", which misses the important visual object ``based ball". This is probably due to the fact that the baseball is very small and the corresponding feature may be lost during processing. Moreover, from the second column of Figure. \ref{example}, we also find that our DSCT can identify more objects (e.g., ``book'', ``tennis'', ``children'', and ``a cup of coffee'') and accurately describes the relationship between objects (e.g., ``near'', ``on the back of'', ``next to''). The above findings indicate that region and segmentation features can be combined to enhance the visual and semantic representations for more accurate and descriptive image captioning.

\subsection{Ablation Studies}
\begin{table}
	\caption{Impact on different feature qualities on COCO Karpathy test split, where B@N, M, R, C and S are short for BLEU@N, METEOR, ROUGE-L, CIDEr and SPICE scores. All values are reported as percentage (\%). }
	\footnotesize
	\begin{center}
		\begin{tabular}{p{2.8cm}|p{0.5cm}p{0.5cm}p{0.5cm}p{0.5cm}p{0.5cm}p{0.5cm}}
			\hline
			Method   &B@1 &B@4  & M &R &C &S\\ \hline  
			Baseline($\rm R_{RES}$)  & 80.9  & 38.7    & 29.1 &58.7 &131.7 &22.8 \\
			Baseline($\rm R_{SEN}$)  & 81.2  & 38.5    & 29.3 &58.9 &131.9 &23.0 \\ \hline
			\textbf{DSCT($\rm R_{RES}$+$\rm S_{Res}$ )} & 82.7  & 40.3    & 30.5 &59.9 &137.6 &23.9   \\ 
			\textbf{DSCT($\rm R_{SEN}$+$\rm S_{Res}$ )} & 82.9  & 40.4    & 30.6 &59.6 &137.9 &23.9   \\ \hline
		\end{tabular}
	\end{center}
	\label{fea}
\end{table}
\textbf{Impact of features. } For the region feature, we separately use two common backbones (ResNet101 \cite{He2016DeepRL} and SENet-154 \cite{Hu2020SqueezeandExcitationN}) to extract the corresponding features, which are denoted by $\rm R_{Res}$ and $\rm R_{SEN}$, respectively. For the segmentation feature, we follow DIFNet \cite{WuDIFNetBV} and use ResNet101 to extract the corresponding features (denoted by $\rm S_{Res}$). We first conduct the experiments by separately using the above two kinds of region features on our baseline transformer, and then we also conduct the experiments by separately integrating one of them and the segmentation feature on our DSCT. The corresponding experimental results are shown in Table. \ref{fea}, from which we can find that the introduction of the segmentation feature improves the captioning performance. Moreover, the performance of image captioning can be further improved by improving the quality of region features.

\begin{table}
	\caption{Impact on different proposed models on COCO Karpathy test split, where B@N, M, R, C and S are short for BLEU@N, METEOR, ROUGE-L, CIDEr and SPICE scores. All values are reported as percentage (\%). }
	\footnotesize
	\begin{center}
		\begin{tabular}{p{2.6cm}|p{0.6cm}p{0.6cm}p{0.5cm}p{0.5cm}p{0.6cm}p{0.5cm}}
			\hline
			Method   &B@1 &B@4  & M &R &C &S\\ \hline  
			Baseline  & 80.9  & 38.7    & 29.1 &58.7 &131.7 &22.8 \\ \hline
			\textbf{PSMAE+} & 81.4  & 39.3    & 29.6 &59.3 &134.1 &23.2 \\
			\textbf{PSMAE++} & 81.7  & 39.6    & 29.7 &59.6 &135.6 &23.3 \\
			\textbf{DSCT(PSMAE+DND)} & 82.7  & 40.3    & 30.5 &59.9 &137.6 &23.9   \\  \hline
		\end{tabular}
	\end{center}
	\label{abl1}
\end{table}
\textbf{Impact of different proposed models. } To comprehensively evaluated our proposed models, we start from a base transformer model (i.e., the baseline), then gradually integrated our PSMAE and DND into it. 
The corresponding experimental results are shown in Table. \ref{abl1}, where \textbf{PSMAE+} means we use the proposed PSMAE to replace the original encoder of the baseline, then we fuse the updated region and segmentation features by element-wise addition operation. \textbf{PSMAE++} means we fuse the updated region and segmentation features by concatenation operation. \textbf{DCST} (i.e., \textbf{PSMAE+DND}) means we replace both the encoder and decoder of the baseline with our proposed PSMAE and DND.

As shown in Table. \ref{abl1}, we can find that \textbf{PSMAE+} and \textbf{PSMAE++} achieve large improvements compared to the baseline, which is mainly caused by 1) the introduction of the segmentation feature helps more accurately describe the spatial and semantic relationships between visual objects, 2) the attention mechanism helps to capture alignment of pattern-specific features for obtaining effective and discriminative representations, and 3) both region and segmentation features are enhanced by querying each other via the proposed PSMAE, which further helps to improve the captioning performance. Then, when we integrate our PSMAE and DND into the baseline, i.e., to construct our DSCT, we can find that the captioning performance is further improved. This indicates that 1) the proposed PSMAE effectively enhances both region and segmentation features and further helps to enhance the visual representation for image captioning. 2) the DND dynamically nominates the right stream at each time to generate the corresponding word in the sentence, which make the generated sentence more descriptive and meantime force each stream in DSCT to focus on learning its private information. 3) by stacking multiple DNDs, the semantic inconsistency and spatial misalignment problem between the region and segmentation features are bypassed and 4) by integrating PSMAE and DND into a novel dual-stream collaborative transformer, our DSCT can utilize the both region and segmentation information more effectively, which helps to achieve the state-of-the-art captioning performance.
\begin{figure*}[t]
	\begin{center}
		\includegraphics[width=0.92\linewidth]{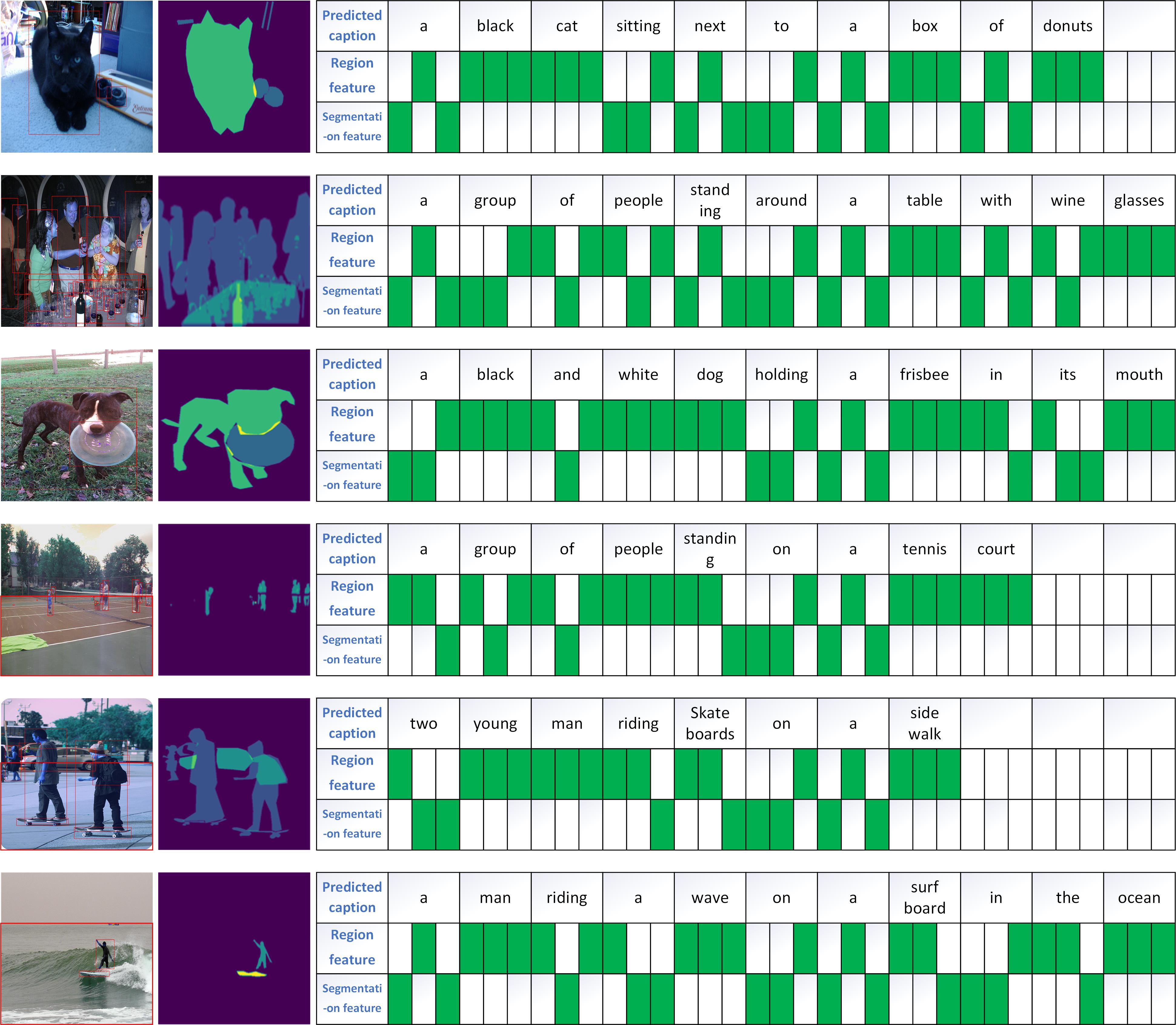}
	\end{center}
	\caption{Visualization of the nominating results corresponding to the proposed Dynamic Nomination Module. The green blocks represent the nominating results. By stacking multiple DNDs, our DSCT effectively fuses region and segmentation features in a dynamic way to generate more accurate and descriptive caption sentences.}
	\label{visual}
\end{figure*}
\begin{table}
	\caption{Impact on different fusion methods on COCO Karpathy test split, where B@N, M, R, C and S are short for BLEU@N, METEOR, ROUGE-L, CIDEr and SPICE scores. All values are reported as percentage (\%). }
	\footnotesize
	\begin{center}
		\begin{tabular}{p{2.8cm}|p{0.5cm}p{0.5cm}p{0.5cm}p{0.5cm}p{0.5cm}p{0.5cm}}
			\hline
			Method   &B@1 &B@4  & M &R &C &S\\ \hline  
			Baseline($\rm R_{Res}$) & 80.9  & 38.7    & 29.1 &58.7 &131.7 &22.8 \\ 
			
			MIA($\rm R_{RES}$+$\rm S_{Res}$)	& 81.7  & 39.2    & 29.6 &59.5 &133.9 &23.3\\
			VSA($\rm R_{RES}$+$\rm S_{Res}$)	& 81.9  & 39.5    & 29.7 &59.7 &134.6 &23.4\\
			IILN($\rm R_{RES}$+$\rm S_{Res}$)  & 81.3  & 38.9    & 29.3 &58.7 &132.7 &23.1\\
				\hline
			\textbf{PSMAE++($\rm R_{RES}$+$\rm S_{Res}$)} & 81.7  & 39.6    & 29.7 &59.6 &135.6 &23.5 \\
			\textbf{DSCT($\rm R_{RES}$+$\rm S_{Res}$)} & 82.7  & 40.3    & 30.5 &59.9 &137.6 &23.9   \\  \hline
		\end{tabular}
	\end{center}
	\label{fusion}
\end{table}

\textbf{Impact of fusion method. } We use different fusion methods (e.g., MIA \cite{Liu2019AligningVR}, VSA \cite{Nagrani2021AttentionBF} and IILN (T=2) \cite{WuDIFNetBV}) to integrate these two pattern-specific features, and the corresponding experimental results are shown in Table. \ref{fusion}. Note that the MIA, VSA and IILN models use the same decoder. As shown in Table. \ref{fusion}, we can find that our DSCT and PSMAE++ outperform the other models, which again indicates our DSCT effectively utilizes the region and segmentation features, thus outperforming the state-of-the-art methods. Note that, compared to MIA, VSA, PSMAE++ and DSCT, IILN obtains the worst results, which is mainly because the core design of IILN is to retain the private information but not the fusion of different pattern-specific features. When changing the original gird and segmentation features into the region and segmentation features, the image captioning performance of IILN is reduced.

\subsection{Self Evaluations}
\textbf{Evaluation of the proposed DNM. }The proposed dynamic nomination module (DNM) in the DND is designed to nominate the appropriate stream to generate each word in sentences, and by stacking multiple DNDs, region and segmentation features are fused more effectively to generate more accurate and descriptive captioning sentences while bypassing their semantic inconsistency and spatial misalignment issues. As our proposed DSCT contains 3 DNDs, we visualize the corresponding nominating results of DNM in each DND in Figure. \ref{visual}. Each word corresponds to six blocks, the green blocks represent the nominating results. As the bean size is set to 5, we select the sentence with the highest CIDEr score for visualization. As shown in Figure. \ref{visual}, we can find that our proposed DND can effectively select the most appropriate stream (i.e., the best pathway) to generate the next word. For example, region features are usually nominated to generate the visual words, such as visual objects (e.g., cat, box, and donuts as shown in the first row of Figure. \ref{visual}) and the attributive of objects (e.g., black, white as shown in the third row of Figure. \ref{visual}). For prepositions expressing spatial relationship (e.g., ``on'' and ``in''), words expressing quantity (e.g., ``a'', ``two'' and ``a group of''), our DND tends to nominate the segmentation stream. The above findings indicate that our DND effectively fuses different pattern-specific features in a dynamic way for helping generate more accurate and descriptive caption sentences.

\subsection{Experimental Results and Discussions}
From the experimental results listed in Tables \ref{taboff}---\ref{fusion} and the figures presented in previous subsections, we have the following observations and corresponding analyses.

(1) UpDown \cite{Anderson2018BottomUpAT}, AoANet \cite{Huang2019AttentionOA}, M2 Transformer \cite{Cornia2020MeshedMemoryTF}, X-Transformer \cite{Pan2020XLinearAN} and our DSCT are all region feature-based image captioning models, while the proposed DCST outperforms UpDown, AoANet, M2 Transformer and X-Transformer as shown in Tables \ref{taboff} and \ref{tabon}. These experimental results indicate that 1) the information of segmentation features helps infer the underlying semantic and spatial relationships among objects for generating more accurate caption sentences. 2) segmentation features are effectively introduced by our proposed DSCT to cooperate with the region features to guide the generation of image captioning sentences, thus more accurate and descriptive image captioning can be achieved by our DSCT. 

(2) DLCT \cite{Luo2021DualLevelCT}, DIFNet \cite{WuDIFNetBV} and our DSCT are all fusion features-based captioning models. Specifically, DLCT improves the image captioning performance by integrating region and grid features, and DIFNet enhances the contribution of visual content for prediction by introducing the segmentation feature. While our proposed DSCT aims to boost image captioning performance by cooperating region features with segmentation features, i.e., we propose a dual-stream collaborative transformer to fuse them. The proposed PSMAE consolidates and updates both region and segmentation features by highlighting their private information and querying each other, and the DND searches and uses the most relevant representations to guide the generation of caption sentences. Therefore, by integrating PSMAE and DND, more effective visual and semantic representations are learned to reduce the dependence on partially generated descriptions, thus achieving more accurate and descriptive image captioning.

(3) Contextual information is very important to image captioning models, especially for region feature-based captioning methods \cite{Anderson2018BottomUpAT, Huang2019AttentionOA, Cornia2020MeshedMemoryTF, Pan2020XLinearAN}. In this paper, we aim to improve the captioning performance by introducing segmentation features then cooperate them with region features. The experimental results as shown in Tables \ref{taboff} and \ref{tabon} show that our proposed DSCT outperforms not only the current region or grid feature-based captioning methods but also fusion feature-based captioning methods, which indicates that our proposed DSCT learns more effective visual and semantic representations to guide the generation of more accurate and descriptive caption sentences.
\section{Conclusion}
Accurate and descriptive image captioning remains a very challenging task due to insufficient visual information and over-reliance on partially generated sentences. In this work, we propose a DSCT model to address these problems by seamlessly integrating the PSMAE and DND models via a two-stream transformer framework.The proposed PSMAE model effectively enhances the representation capability of features in each stream, and the DND model is able to choose the proper stream to generate each word in the caption sentences, which in turn forces each stream in PSMAE to learn and update its private information. Hence, by integrating PSMAE and DND, more effective visual and semantic representations are learned to reduce the dependence on partially generated descriptions for predicting the remaining words. This leads our DSCT achieving more accurate and descriptive image captioning than the state-of-the-art models. It can also be found from the experiments that by cooperating the region feature and segmentation feature in a dynamic way, their semantic inconsistency and spatial misalignment issues are bypassed and their homogeneous features are effectively mined to improve the captioning performance. In the future, we plan to conduct research on endowing human-like controllability to the captioning model and extend our model to other related topics.


%


\ifCLASSOPTIONcaptionsoff
  \newpage
\fi


%



\bibliographystyle{IEEEtran}
\bibliography{IEEEabrv,IEEEexample}

\end{document}